\documentclass[a4paper,fleqn]{cas-dc}

\usepackage[backend=biber, style=numeric]{biblatex}
\usepackage{subfigure}
\def\tsc#1{\csdef{#1}{\textsc{\lowercase{#1}}\xspace}}
\tsc{WGM}
\tsc{QE}
\tsc{EP}
\tsc{PMS}
\tsc{BEC}
\tsc{DE}

\begin{document}
\let\WriteBookmarks\relax
\def\floatpagepagefraction{1}
\def\textpagefraction{.001}
\shorttitle{PatchDenoiser}
\shortauthors{Jitindra Fartiyal et~al.}

\title [mode = title]{PatchDenoiser: Parameter-efficient multi-scale patch learning and fusion denoiser for Low-dose CT imaging}                      
\tnotemark[1,2]

\tnotetext[1]{This research was funded by the  European Union's Horizon Europe project BETTER (grant agreement No 101136262). SKT acknowledges support of the UK Multidisciplinary Centre for Neuromorphic Computing (UKRI982).}

\author[1]{Jitindra Fartiyal}[type=editor, orcid=0009-0002-2937-6258]
\ead{240304225@aston.ac.uk}
\affiliation[1]{organization={Aston Institute of Photonic Technologies, Aston University},
                addressline={Aston Triangle}, 
                city={Birmingham},
                postcode={B4 7ET}, 
                state={West Midlands},
                country={United Kingdom}}

\author[1]{Pedro Freire}
\ead{p.freiredecarvalhosourza@aston.ac.uk}

\author[1]{Sergei K. Turitsyn}
\ead{s.k.turitsyn@aston.ac.uk}

\author[1]{Sergei G. Solovski}
\ead{s.sokolovsky@aston.ac.uk}

\cortext[cor1]{Corresponding author}
\cortext[cor2]{Principal corresponding author}

\begin{abstract}
Low-dose computed tomography (LDCT) plays a critical role in cancer screening, pediatric imaging, and longitudinal monitoring, where repeated imaging is required while minimizing radiation exposure. Despite its clinical importance, the acquisition of LDCT often introduces substantial noise and artifacts that can obscure anatomical structures, reduce diagnostic confidence, and negatively impact downstream image analysis tasks. Traditional denoising approaches based on handcrafted filtering techniques frequently over-smooth anatomical details, whereas recent deep learning methods, including CNN, GAN, and transformer-based architectures, often rely on large-scale models with high computational and energy demands, limiting their practicality for resource-constrained clinical deployment.

In this work, we propose PatchDenoiser, a lightweight and energy-efficient patch-based denoising framework for LDCT restoration. The proposed architecture decomposes denoising into complementary stages: local texture extraction and global contextual aggregation, followed by a spatially aware patch fusion mechanism to preserve fine anatomical structures while effectively suppressing noise. Unlike existing large-capacity denoising networks, PatchDenoiser is specifically designed for computational efficiency, substantially reducing model complexity while maintaining competitive reconstruction quality.

Extensive experiments on four LDCT datasets demonstrate that PatchDenoiser achieves comparable quantitative and perceptual performance compared to state-of-the-art CNN-based denoising methods. In particular, the proposed framework requires approximately 10$\times$ fewer parameters and 38$\times$ lower computational complexity than conventional CNN-based approaches, while maintaining comparable PSNR and SSIM performance. These results highlight the strong efficiency-performance trade-off achieved by PatchDenoiser and demonstrate its suitability for real-world clinical deployment and scalable medical imaging pipelines.

Overall, PatchDenoiser provides an effective balance between reconstruction fidelity, computational efficiency, and deployment practicality, making it a promising solution for efficient LDCT denoising. The source code is publicly available at: https://github.com/JitindraFartiyal/PatchDenoiser.
\end{abstract}


\maketitle

\section{Introduction}

Medical images play a vital role in disease diagnosis, health monitoring, treatment planning, and a wide range of clinical applications. They also form the foundation of many artificial intelligence(AI) based downstream tasks, including disease detection, anatomical and pathological segmentation, and broader healthcare research. However, medical images are frequently degraded or corrupted by noise arising from factors such as low-quality or low-dose acquisition protocols, patient motion, hardware constraints, and inherent scanner limitations. The presence of noise can impede accurate diagnosis and treatment planning and can significantly degrade the performance of downstream image analysis algorithms \cite{Ferdinand2022, Vannadil2023}. Therefore, effective noise removal is a critical preprocessing step in medical image analysis pipelines.

Noise removal is particularly important in computed tomography (CT). CT scans expose patients to ionizing radiation, which may pose long-term health risks \cite{Pearce2012}. To reduce radiation exposure, low-dose CT (LDCT) acquisition protocols are commonly adopted. Although LDCT improves patient safety, it introduces substantial noise into the images, making clinical interpretation and diagnosis more challenging \cite{Ehman2014}. Consequently, effective denoising methods are crucial for enhancing LDCT image quality while preserving clinically relevant structures.

Traditional noise removal techniques for CT images, such as filtering-based methods, can suppress noise but often over-smooth medical images, leading to the loss of fine anatomical details \cite{Kaur2021, Kaur2023, Abuya2023}. With the recent development of convolutional neural networks (CNNs), CNN-based denoisers have demonstrated improved denoising performance with reduced smoothing compared to traditional approaches \cite{Singh2022, Sadia2024}. However, CNN-based methods can still produce over-smoothed patches, which reduces the visibility of small vessels and subtle tissue structures \cite{Wenchao2024}. More recently, transformer-based and generative adversarial network (GAN)-based denoisers have been proposed to alleviate excessive smoothing and improve perceptual quality \cite{Jameel2025, Shikai2025, Kaur2023}. Despite these advances, such models are typically large and computationally expensive, requiring substantial training data and extensive fine-tuning \cite{Fang2025, Willemink2022, Tavakoli2025, Nazir2024}. Furthermore, recent studies report limited generalizability across different scanners, acquisition protocols, and clinical conditions. In summary, existing approaches face a trade-off between relatively lightweight CNN-based models with limited performance and highly complex transformer or GAN-based models with large parameter counts and data requirements.

\begin{figure}
    \centering
    \begin{subfigure}
        \centering
         \includegraphics[width=0.5\textwidth]{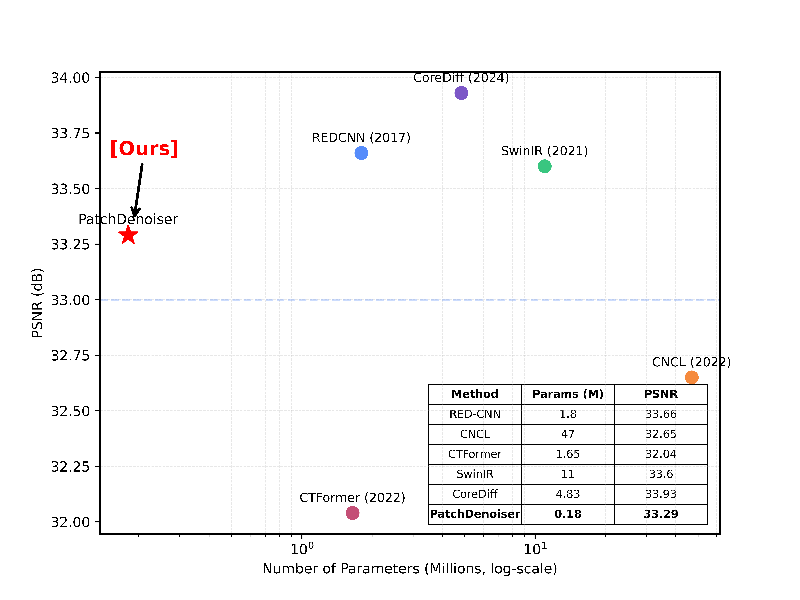}
    \end{subfigure}
    \begin{subfigure}
        \centering
         \includegraphics[width=0.5\textwidth]{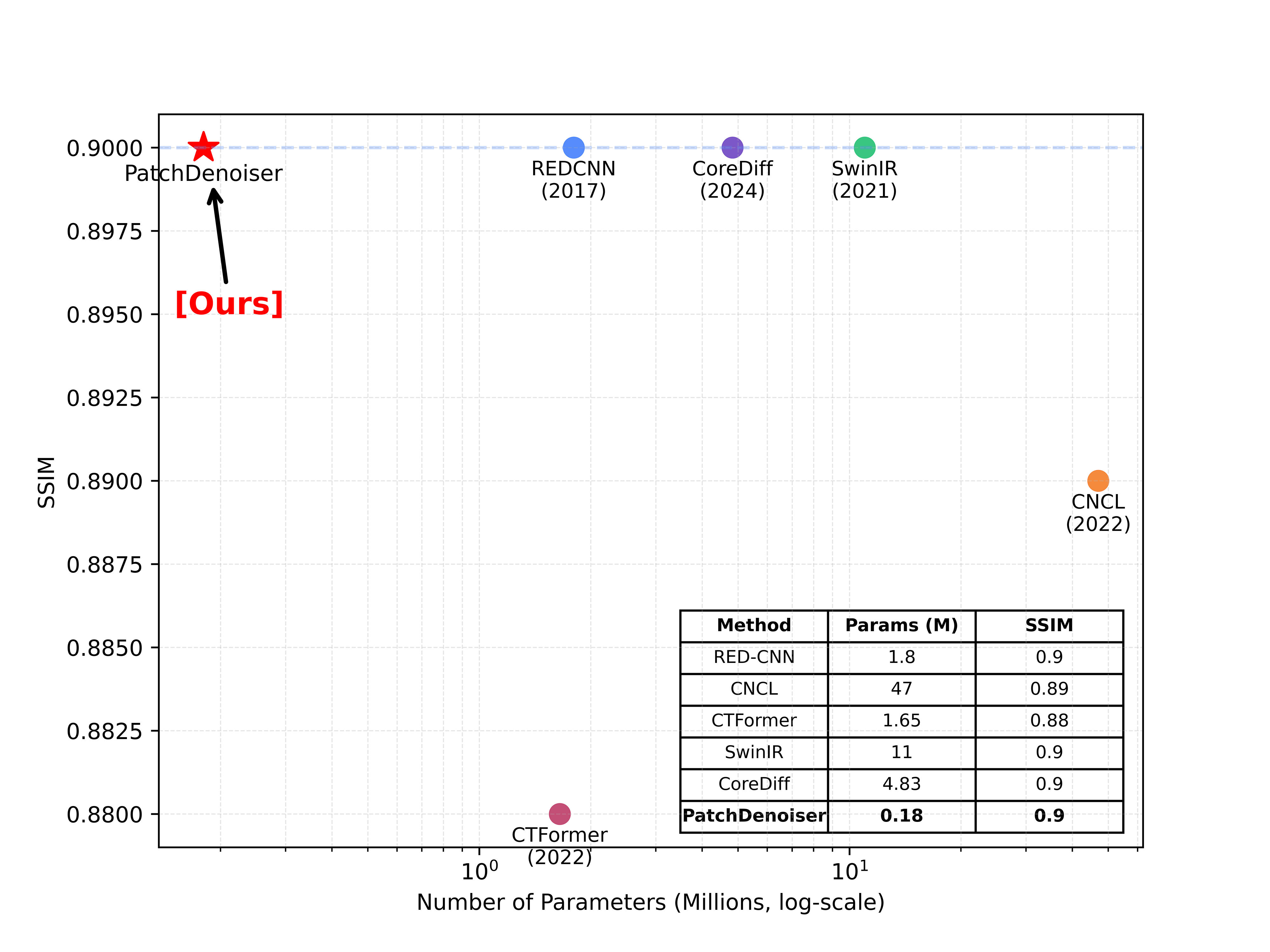}
    \end{subfigure}
    \caption{\textbf{Trade-off between model complexity and reconstruction quality.} Top: Peak Signal-to-Noise Ratio (PSNR) vs model parameters, illustrating the performance–complexity trade-off across different algorithms. Bottom: Structural Similarity Index Measure (SSIM) vs model parameters, showing consistency of perceptual quality improvements across different algorithms. Higher PSNR/SSIM values indicate better reconstruction quality, while fewer parameters correspond to lower model complexity.}
    \label{fig:psnr_ssim_parameter_trade_off}
\end{figure}

At the same time, research on deep learning architectures is approaching a point where increased model complexity leads to disproportionately high energy consumption \cite{Prajwal2025, Rozycki2025, Liu2022}. For denoising tasks that often serve as a preprocessing component within larger AI-based medical analysis pipelines, models must be lightweight and energy-efficient while still achieving high denoising performance. To address these challenges, we propose a new CNN-based denoising architecture based on patch-based learning. The proposed model is ultra-lightweight, with approximately 10x fewer parameters and 38x lower compute cost than traditional CNN-based architectures, while achieving competitive performance in peak signal-to-noise ratio (PSNR) and structural similarity index (SSIM) for CT image denoising. By providing high-quality denoised CT images with minimal computational and energy requirements, our model can facilitate more accurate diagnoses and streamlined clinical workflows, particularly in settings with limited computational resources. Figure \ref{fig:psnr_ssim_parameter_trade_off} presents the PSNR-Parameter and SSIM-Parameter trade-off plots, which clearly show that PatchDenoiser has the best PSNR and SSIM parameter efficiency.

The main contributions of this work are as follows:
\begin{itemize}
   \item We design a parameter-efficient denoising model that, to the best of our knowledge, is the smallest    model achieving competitive performance with the lowest carbon emissions under identical hardware settings.
    \item We propose a multi-scale patch-based denoising framework that learns from individual patches while preserving spatial information.
    \item We introduce an efficient spatial-aware patch fusion strategy that integrates information from different scale patches.
    \item We evaluated our models extensively on four different LDCT datasets with three different anatomical regions and two different Hounsfield Unit (HU) window.
\end{itemize}

\section{Related Works}
Before the deep learning revolution, window-based filtering denoisers were widely used. These methods include mean, median, and Gaussian filtering, in which the average replaces each pixel or a statistically computed value within an N×N window \cite{Irum2015, Yu2025}. Such approaches are simple and computationally efficient. However, window-based averaging leads to excessive smoothing and loss of fine structural details \cite{Ullah2025, Jain2016}. To capture non-local information in images, the non-local means (NLM) algorithm \cite{Buades2011} was introduced, which extends beyond the local neighborhood by identifying and averaging similar patches across the image. Block-matching and 3D filtering (BM3D) \cite{Dabov2007} further improves upon this idea by grouping similar blocks into 3D volumes and performing collaborative filtering in the transform domain, followed by inverse transformation. Although NLM and BM3D outperform window-based filtering methods, they still suffer from smoothing artifacts, increased algorithmic complexity, and a limited ability to preserve global structural details accurately.

Compared with traditional algorithms, deep learning–based methods have demonstrated substantial performance improvements, particularly in capturing global structures and complex anatomical patterns. For example, RED-CNN \cite{Chen2017} is a popular CNN-based denoising model that adopts a symmetric autoencoder architecture with residual shortcut connections to improve feature propagation. The Dynamic Residual Attention Network (DRAN), introduced by Sharif et al. \cite{Sharif2020}, incorporates dynamic convolution operations with spatial attention mechanisms to reduce excessive smoothing. Furthermore, Sharif et al. \cite{Sharif2024} proposed a cascaded two-stage framework that integrates residual dense blocks with a noise-aware attention module for more accurate pixel-wise noise estimation. In the context of ultrasound imaging, the feature-driven convolutional network (FDCCN) \cite{Dong2021} exploits guided backpropagation and dilated convolutions to enhance despeckling performance. To strengthen multi-scale feature representation, Yang et al. \cite{Yang2023} proposed a dual-network strategy consisting of a Feature Refinement Network (FRN) and a Dynamic Perception Network (DPN). Recent advances have focused on enhancing receptive field capabilities and feature representation in CNN-based denoising architectures. The integration of dilated convolutions with preprocessing and post-processing techniques has enabled larger receptive fields, allowing distant pixels to contribute to feature map enrichment for more effective LDCT denoising \cite{Trung_DCNNERF_2022}. To overcome the limitations of fixed-size convolutional kernels, novel approaches have emerged, such as the quadratic autoencoder \cite{Fan_QAE_2020}, which replaces traditional inner products with quadratic operations to enhance individual neurons' capabilities and model efficiency.

Furthermore, self-attention mechanisms have been successfully implemented in 3D CNN architectures \cite{Li_SACNN_2020}, enabling the capture of spatial relationships within and between slices while leveraging self-supervised perceptual loss functions to improve performance. Overall, CNN-based denoising algorithms outperform traditional methods by better preserving structural integrity. However, despite reduced smoothing compared to classical approaches, denoised images can still exhibit excessive smoothness and loss of high-frequency details.

To further mitigate smoothing artifacts, several generative adversarial network (GAN)-based and transformer-based methods have been proposed. One notable work is CNCL by Geng et al. \cite{Geng2022}, which introduces separate models for the noise and content components, which are fused and evaluated with a PatchGAN discriminator, thereby avoiding conventional L1/L2-driven objectives. Extending this dual-signal concept, Huang et al. \cite{Huang2022} proposed a dual-domain GAN architecture that operates in both the image and gradient spaces to better preserve edges and fine details. Adversarial and structural loss functions have also been explored, including the GAN-based ultrasound despeckling method proposed by Mishra et al. \cite{Mishra2018}, which mitigates the smoothing introduced by conventional loss functions while improving visual fidelity. Recent advances have also focused on specialized detail recovery and domain knowledge integration for medical image denoising. SDGAN \cite{Wang_DRMID_2019} demonstrates improved detail reconstruction through a two-stage approach that first performs preliminary denoising before specifically recovering lost textures, showing significant gains in PSNR and SSIM metrics for Wireless Capsule Endoscopic images. Multi-task learning is also used, as in MTD-GAN \cite{Kyung_MTDGAN_2022}, which introduces a multi-task discriminator framework that simultaneously performs classification, segmentation, and reconstruction to preserve fine details. Along similar lines, recent work by \cite{Yin_SDKGLDCT_2022} has successfully leveraged domain knowledge through segmentation-guided approaches, demonstrating that integrating precise semantic segmentation information can significantly enhance LDCT denoising performance and enable coarse-grained knowledge sharing across different datasets. Despite these improvements, GAN-based models remain challenging to train, are computationally expensive, and are prone to generating hallucinated or novel artifacts.

More recently, transformer-based denoising models have been investigated. CSFormer \cite{Yin2022} enhances denoising performance by integrating cross-scale feature fusion with Swin Transformer blocks, enabling the modeling of both local textures and long-range dependencies. Another transformer-based model, CTFormer \cite{Wang2023}, employs token-rearrangement strategies to encode local texture information and global contextual interactions jointly. Compared to GAN-based methods, transformer-based models are less prone to introducing novel artifacts and generally achieve improved visual fidelity with reduced smoothing. Transformer architectures like Hformer \cite{Zhang_Hformer_2023} have focused on improving computational efficiency and local feature capture by combining CNN-based local extraction with transformer-based global modeling through lightweight depthwise separable convolutions and efficient cross-covariance attention. Along similar lines, CT-Denoimer \cite{Zhang_CTDenoimer_2025} introduces an efficient framework that captures both global correlations through the multi-Dconv head transposed attention (MDTA) module with linear computational complexity and intricate spatially varying local contextual details via the mixed contextual feed-forward network (MCFN). To further enhance collaborative refinement, incorporating operation-wise attention layers enables more effective handling of complex and varied noise patterns in LDCT images. However, transformer-based approaches typically require large-scale training datasets, are susceptible to overfitting, and incur high computational and training costs.

In summary, deep learning–based denoising models outperform traditional algorithms in both noise suppression and structural preservation. Among these approaches, CNN-based models are simpler and faster to train but remain susceptible to over-smoothing. In contrast, more complex GAN-based and transformer-based models alleviate smoothing at the cost of increased model complexity and computational overhead. Furthermore, fair and accurate comparison among existing methods remains challenging due to the lack of standardized evaluation datasets and benchmarking protocols. Several studies rely on private datasets, limiting reproducibility, while others perform limited evaluations across acquisition parameters, making it difficult to assess generalizability. Consequently, the performance gains reported by GAN- and transformer-based approaches may be dataset- or model-specific.

These limitations motivate the need for a denoising framework that combines the efficiency of CNN-based models with improved multi-scale representation and reduced smoothing, without incurring the computational overhead of GAN or transformer-based approaches.

\section{Method}
PatchDenoiser is a CNN-based, spatially aware, multi-scale denoising architecture designed to achieve high performance with fewer parameters. Although image denoising is primarily a local operation, global interactions still play an important role in preserving structural consistency. Conventional CNN-based models, particularly those trained on patches, often fail to capture such global interactions adequately. PatchDenoiser addresses this limitation by structurally separating local texture extraction from global feature integration. The architecture captures local textures through patch-level processing and subsequently fuses information across scales to model global interactions.

\begin{figure*}
    \centering
    \includegraphics[width=\textwidth]{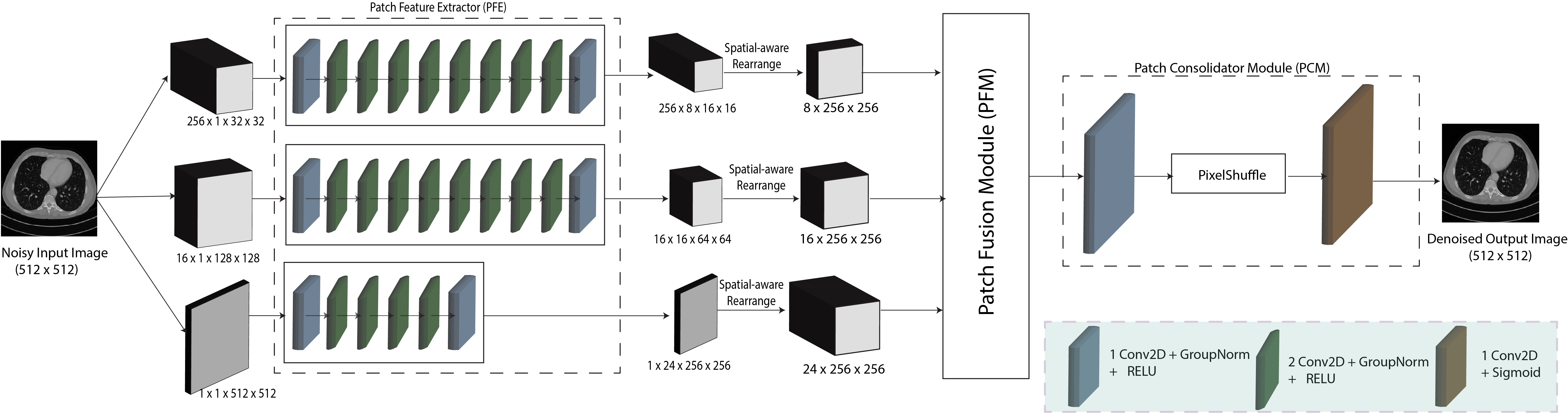}
    \caption{Overview of the PatchDenoiser framework showing the three main modules: Patch Feature Extractor (PFE), Patch Fusion Module (PFM), and Patch Consolidator Module (PCM).}
    \label{fig1}
\end{figure*}

PatchDenoiser consists of three main modules: the Patch Feature Extractor (PFE), the Patch Fusion Module (PFM), and the Patch Consolidator Module (PCM). The PFE extracts noise-free content from image patches at multiple scales. The PFM fuses information from lower-scale patches to higher-scale patches in a spatially aware manner, enabling effective cross-scale information flow. And, the PCM module consolidates the fused feature maps and removes patch boundary artifacts to produce the final denoised output. Figure \ref{fig1} illustrates the overall PatchDenoiser architecture.

\subsection{Patch Feature Extractor (PFE)}
At the input stage, patches of multiple scales are generated from the image. To capture both small and large-scale features, patch sizes are determined by dividing the image resolution by 16, 8, and 1. For example, for an input image of size 512×512, patch sizes of 32×32, 128×128, and 512×512 are used. Instead of employing a single PFE configuration across all patch sizes, PatchDenoiser uses PFEs with varying depths and latent dimensions.

For small patch sizes (e.g., 32×32), a deeper PFE with a smaller latent dimension is used. Since small patches contain limited contextual information, increased depth helps extract meaningful representations, whereas a reduced latent dimension is sufficient given their lower information content. In contrast, for large patch sizes (e.g., 512×512), a shallower PFE with a larger latent dimension is employed to capture global contextual information better.

In addition, different kernel sizes are used in the initial layer for different patch scales: 3, 7, and 11 for patch sizes of 32, 128, and 512, respectively. Larger kernels for larger patch sizes enable more effective capture of global features. Unlike conventional autoencoder architectures, the spatial resolution is not aggressively reduced; instead, the spatial dimension is preserved at half of the input resolution at the output of the PFE. This structurally defined design helps minimize information loss while achieving parameter efficiency.

\subsection{Patch Fusion Module (PFM)}
Following the PFE stage, feature maps corresponding to different patch scales are obtained. These multi-scale feature maps are then fused using a spatially aware strategy, in which lower-scale maps are integrated with higher-scale maps based on their spatial positions. This design ensures proper alignment and information flow across scales.

\begin{figure*}
    \centering
    \includegraphics[width=\textwidth]{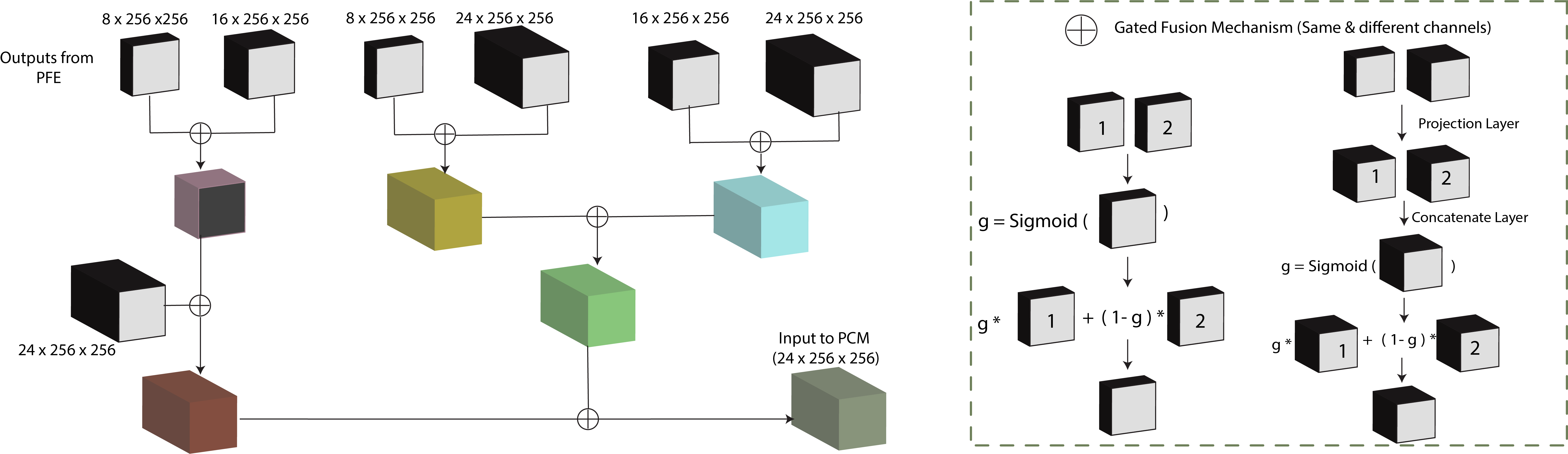}
    \caption{Illustration of the Patch Fusion Module (PFM), where multi-scale feature maps are combined using a gated fusion mechanism.}
    \label{fig2}
\end{figure*}

Fusion is performed using a gated fusion mechanism, where a sigmoid activation function acts as a gate to regulate the contribution of each feature map. Fusion occurs across all PFE outputs to produce a single fused representation. The resulting fused feature map is then passed to the Patch Consolidator Module. The structural fusion process implemented in the PFM is illustrated in Figure \ref{fig2}.

\subsection{Patch Consolidator Module (PCM)}
The fused output produced by the PFM already contains denoised image content; however, due to patch-based processing, residual patch boundary artifacts may still be present. The Patch Consolidator Module is a lightweight module composed of convolutional and upsampling layers designed to remove these artifacts. By refining spatial continuity across patch boundaries, the PCM produces the final denoised output image.

\section{Experimental Setup}
\subsection{Datasets}
For evaluation, we use four publicly available datasets. These datasets cover three anatomical regions: the abdomen, the chest, and an entire piglet. The evaluation dataset comprises two Hounsfield unit (HU) windows: (-1000, 400) for chest images and (-160, 240) for the rest. It also covers small-scale to medium-scale datasets for evaluation.

\subsubsection{Mayo 2016 Dataset}
We are using the 2016 Low Dose CT Grand Challenge (shortened as \textit{mayo2016ABD}) \cite{McCollough2017}, which contains low-dose (25\%) and full-dose abdomen CT scans for 10 patients. It provides four configurations with varying kernel and slice thicknesses. For our primary comparison, we use the B30 kernel with a 3mm slice thickness, as it is the most common and balanced kernel and is supported by prior work \cite{Chen2017, Geng2022, Li_SACNN_2020}. It contains 512 x 512 2,378 abdomen CT scan images. For our training and validation, we used a 7:3 subject split and uniformly extracted 100 slices from each patient, resulting in 700 and 300 scans for the training and validation phases, respectively. For preprocessing, we followed a common HU window for the abdomen \cite{Murphy2026, Hartung2026} and employed a [-160, 240] HU window and normalized it to [0, 1]. 

\subsubsection{Mayo 2020 Dataset}
We also employed the TCIA Cancer Archive Mayo 2020 dataset [7] to organize two datasets. One for Low dose (10\%), and Full dose chest CT scans for 25 patients (shortened as \textit{mayo2020CHEST}) and one for Low dose (25\%), and Full dose abdomen CT scans for 25 patients (shortened as \textit{mayo2020ABD}). A note that the abdominal CT scans for 25 patients differ from the Mayo LDCT 2016 dataset. The Mayo2020CHEST dataset contains 512 x 512 chest CT scans (8662). For our evaluation, we selected 20 random patients for training and the remaining 5 for validation, uniformly extracting 100 slices from each patient, resulting in 2000:500 scans for the training and validation phases. For preprocessing, we followed the guidelines in \cite{CTRadCafe2026} and applied a HU window of [-1000, 400], then normalized the data to [0, 1]. The Mayo2020ABD dataset contains 512 x 512 4228 abdomen CT scans. For our evaluation, we selected 20 random patients for training and the remaining 5 for validation, uniformly extracting 100 slices from each patient, resulting in 1990:500 scans for the training and validation phases. For preprocessing, we followed the guidelines in \cite{Murphy2026, Hartung2026} and applied a [-160, 240] HU window, then normalized the data to [0,1].

\subsubsection{Piglet dataset}
We also employed the Piglet dataset (shortened as \textit{piglet}) \cite{Yi_Babyn_2018}, which contains 850 CT scans from a GE Discovery CT750 HD scanner of real piglets with different doses. For our evaluation, we used a pair of 10\% dose and normal dose pairs, as 10\% dose samples would be hard to denoise and would be a strong benchmark point to evaluate.

\subsection{Preprocessing}
CT images contain raw pixel intensity values that depend on scanner-specific acquisition parameters. To obtain a standardized and scanner-independent representation, the raw pixel values are converted to Hounsfield Units (HU) using the formula \eqref{eq1}.

\begin{equation}
    \label{eq1}
    HUImg = Img * RescaleSlope + RescaleIntercept
\end{equation}

This preprocessing strategy is applied before HU clipping and normalization and consistently for both training and testing across all evaluated methods and datasets to ensure a fair comparison.

\subsection{Training}
All models are implemented using PyTorch. We employ the L2 loss function \eqref{eq1} and optimize the network parameters using the AdamW optimizer with a cosine annealing learning rate schedule. The initial learning rate is set to $1e^{-2}$, with a minimum learning rate $\eta_{min} = 1e^{-6}$. Training is performed with a batch size of 16 on an NVIDIA RTX 5000 Ada Generation GPU. 

\begin{equation}
    \label{eq1}
    Loss = \sum_{i=1}^N (y_{gt} - y_{pred})^2
\end{equation}

\subsection{Evaluation Design}
For our primary evaluation, we compare the performance of PatchDenoiser against traditional SOTA algorithms like BM3D, CNN-based SOTA algorithms like RED-CNN, GAN-based SOTA algorithms like CNCL, transformer-based SOTA algorithms like SwinIR and CTFormer, and diffusion-based algorithms like CoreDiff. To ensure a fair comparison across architectures with varying computational complexity, we adopt a compute-aware unified training protocol across all methods. All models are trained on identical train/validation splits for each dataset, including \textit{mayo2016ABD}, \textit{mayo2020ABD}, \textit{mayo2020CHEST}, and \textit{piglet} datasets. The training budget is allocated in terms of optimization iterations, scaled proportionally with dataset size to ensure balanced exposure across datasets. Since convergence behavior varies across architectures (CNN, GAN, Transformer, and Diffusion-based models), we further allow architecture-dependent iteration ceilings to account for differences in convergence speed. We also employ an early stopping strategy based on validation PSNR, where training is terminated if the average improvement in PSNR over three consecutive validation evaluations is less than 0.05 dB. Model checkpoints are saved at every epoch (or equivalent iteration interval) to ensure selection of the best-performing model. Since different architectures exhibit substantially different convergence rates, a fixed iteration budget may bias the comparison toward faster-converging models. Therefore, we adopt a compute-aware protocol where each method is allowed sufficient optimization steps to reach convergence, with early stopping based on validation performance ensuring that no model benefits from unnecessary additional training. Table \ref{tab:training_details} presents the comprehensive details of the compute-aware unified training protocol displayed after the bibliography.

A note of CoreDiff evaluation, which needs consecutive slices of scans, is removed from the piglet dataset evaluation as the scans are not consecutive in nature.

This strategy provides a consistent yet compute-efficient training framework while avoiding overfitting and excessive training redundancy across architectures with heterogeneous convergence characteristics.

\subsection{Evaluation Metrics}
For quantitative analysis, we have used three metrics: Peak Signal-to-Noise Ratio (PSNR) and Structural Similarity Index Measure (SSIM). PSNR measures noise reduction, whereas SSIM is a perceptual metric that assesses degradation in image visual quality. In addition, following the work of \cite{Gao_Li_Zhang_Zhang_Shan_2024}, we have also used Feature Similarity Index (FSIM) and Noise Quality Metric (NQM), which can correlate to some degree with the subjective assessments of image quality by the radiologist \cite{Mason_Rioux_Clarke_Costa_Schmidt_Keough_Huynh_Beyea_2020}. All the metrics are calculated after renormalizing the output back to their respective HU window.
For Qualitative analysis, we selected a 128x128 window from the entire validation dataset where the Normal-dose CT (NDCT) and Low-dose CT (LDCT) have a large error difference and gradient, signifying a high-noise, information-rich region of interest, and randomly selected from the top 1000 slices.

\subsection{Computational Efficiency and Environmental Metrics}
We evaluate the computational efficiency and environmental impact of all methods using a comprehensive set of metrics. Specifically, we report training time (in hours) on the \textit{mayo2020ABD} dataset and measure single-sample GPU inference latency (ms) under identical hardware settings. Model complexity is quantified by the number of parameters (in millions at test time) and the computational cost, measured in GFLOPs (at test time), using the fvcore library. In addition, we assess the environmental footprint by estimating carbon emissions (kg $CO_2$) using CodeCarbon. Together, these metrics provide a balanced evaluation of accuracy, efficiency, and sustainability, enabling fair comparison across competing methods.

\section{Results}
\subsection{Computational Efficiency and Resource Footprint}
We first analyze the computational efficiency and resource footprint of PatchDenoiser. As shown in Table \ref{tab:computational-efficiency-table}, the proposed method consistently reduces model complexity in terms of parameters and FLOPs, while achieving lower inference latency and training cost across all datasets, and maintaining competitive PSNR among state-of-the-art algorithms. In addition, we quantify environmental impact using estimated carbon emissions, demonstrating a substantially reduced footprint compared to competing methods. A special exception is for SwinIR, where the image size is 128x128 due to low GPU memory; the rest of the algorithms are evaluated on the same hardware and image settings.

\begin{table*}
\centering
\caption{Comprehensive comparison of reconstruction quality and computational efficiency across datasets. PSNR values are reported as mean $\pm$ std.}
\label{tab:computational-efficiency-table}
\resizebox{\textwidth}{!}{
\begin{tabular}{l|c c c c|c c c c c}
\hline
\multirow{2}{*}{Models} 
& \multicolumn{4}{c|}{PSNR (dB)} 
& \multirow{2}{*}{Train Time (h)} 
& \multirow{2}{*}{Inference (ms)} 
& \multirow{2}{*}{Params (M)} 
& \multirow{2}{*}{FLOPs (G)} 
& \multirow{2}{*}{CO$_2$ (kg)} \\
\cline{2-5}
& Mayo2016ABD & Mayo2020ABD & Mayo2020Chest & Piglet & & & & & \\
\hline

LDCT        & 27.72$\pm$2.20 & 29.17$\pm$2.70 & 13.48$\pm$2.24 & 29.35$\pm$4.13 & NA & NA & NA & NA & NA \\
BM3D        & 29.44$\pm$2.53 & 31.28$\pm$3.09 & 13.56$\pm$2.26 & 30.18$\pm$4.11 & NA & NA & NA & NA & NA \\
RED-CNN     & 32.16$\pm$1.75 & 33.67$\pm$2.33 & 20.62$\pm$2.42 & 34.90$\pm$3.73 & 0.40 & 27 & 1.80 & 458 & 2.78e-4 \\
CNCL        & 31.12$\pm$1.43 & 32.65$\pm$2.05 & 20.47$\pm$2.33 & 33.45$\pm$3.22 & 0.60 & 18 & 47 & 291 & 3.01e-4 \\
CTFormer    & 30.88$\pm$1.76 & 32.05$\pm$2.41 & 19.45$\pm$2.23 & 32.11$\pm$3.31 & 0.45 & 140 & 1.65 & 1361 & 1.49e-3 \\
SwinIR      & 32.51$\pm$1.82 & 33.60$\pm$2.51 & 20.72$\pm$2.48 & 32.53$\pm$3.38 & 3.40 & 53 & 11 & 201 & 1.49e-2 \\
CoreDiff    & 32.38$\pm$1.84 & 33.93$\pm$2.18 & 20.77$\pm$2.52 & NA & 3.66 & 104 & 4.83 & 151 & 1.06e-3 \\
PatchDenoiser & 32.02$\pm$1.67 & 33.30$\pm$2.24 & 20.64$\pm$2.44 & 34.77$\pm$3.47 & \textbf{0.80} & \textbf{4.2} & \textbf{0.18} & \textbf{12} & \textbf{7.74e-5} \\
\hline
\end{tabular}}
\end{table*}

\begin{figure}
    \centering
    \includegraphics[width=0.5\textwidth]{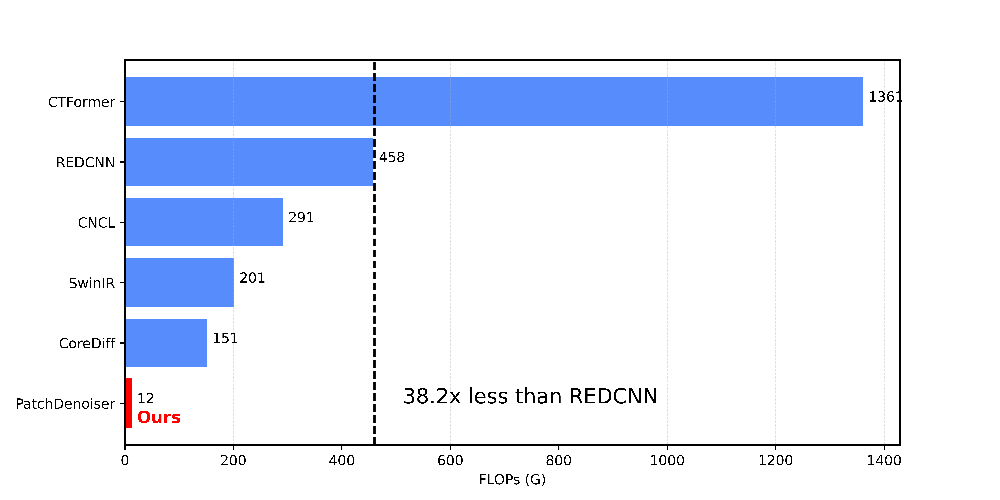}
    \caption{Computational complexity comparison in terms of FLOPs across different methods.}
    \label{fig:flops_efficiency}
\end{figure}

\begin{figure}
    \centering
    \includegraphics[width=0.5\textwidth]{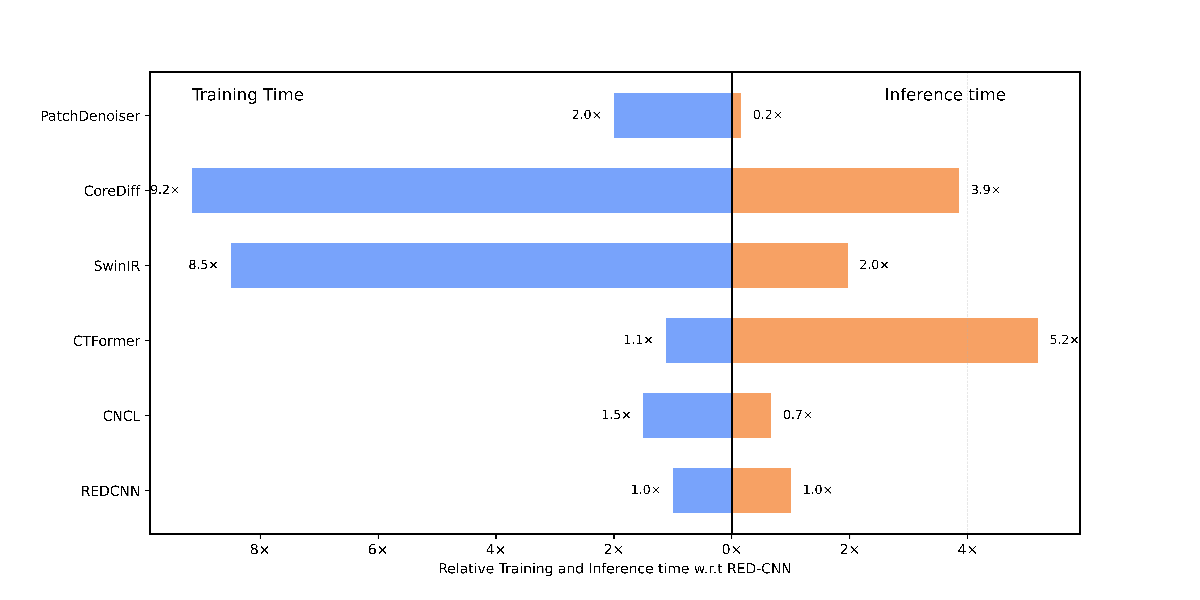}
    \caption{Normalized training and inference time comparison with RED-CNN as the reference baseline.}
    \label{fig:training_efficiency}
\end{figure}

\begin{figure}
    \centering
    \includegraphics[width=0.5\textwidth]{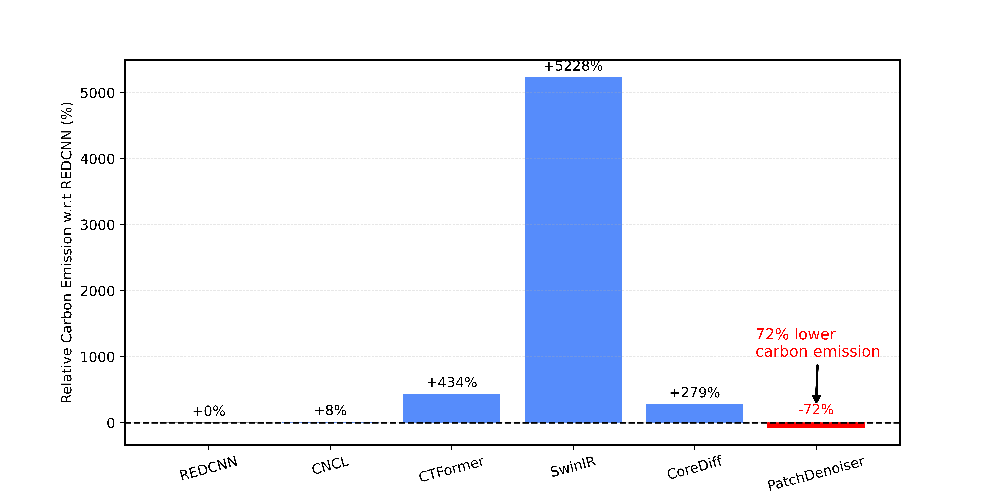}
    \caption{Environmental impact (carbon emission) comparison measured using CodeCarbon with RED-CNN as the reference baseline.}
    \label{fig:carbon_emission}
\end{figure}

To further analyze the computational advantages of PatchDenoiser, we present a detailed efficiency study of FLOPs, training and inference times, and carbon emissions. As shown in Figure \ref{fig:flops_efficiency}, PatchDenoiser requires approximately 38$\times$ fewer FLOPs compared to RED-CNN, highlighting its significantly reduced computational complexity. We further evaluate runtime efficiency by normalizing training and inference times with respect to RED-CNN. As illustrated in Figure \ref{fig:training_efficiency}, although PatchDenoiser has a modest increase in training time due to a multi-scale patch processing strategy, it achieves substantially faster inference with nearly 0.2$\times$ the runtime of RED-CNN. Moreover, compared to diffusion-based approaches, PatchDenoiser demonstrates markedly improved deployment efficiency, requiring up to 24$\times$ less inference time. Finally, we assess environmental impact using CodeCarbon-based estimates. As shown in Figure \ref{fig:carbon_emission}, PatchDenoiser achieves a 72\% reduction in carbon emissions relative to RED-CNN, emphasizing its suitability for sustainable and energy-efficient medical imaging applications.

\subsection{Perceptual Quality Analysis}
We further evaluate the perceptual reconstruction quality of the proposed PatchDenoiser using both quantitative metrics and qualitative visual assessment. As reported in Table \ref{tab:perceptual_metrics}, PatchDenoiser consistently achieves competitive or superior performance in perceptual similarity measures, including FSIM and NQM, across all datasets, indicating improved structural fidelity and better alignment with visually perceived image quality. These results demonstrate that the proposed method not only improves computational efficiency but also maintains high-quality reconstruction performance. To complement the quantitative analysis, we present qualitative comparisons across all four datasets using representative top-performing methods, including RED-CNN, SwinIR, CoreDiff, and the proposed PatchDenoiser. The visual results further validate that PatchDenoiser effectively suppresses noise while preserving fine anatomical structures, yielding sharper and more visually consistent reconstructions.

\begin{table*}
\centering
\caption{Comparison of perceptual metrics across different datasets and denoising methods.}
\label{tab:perceptual_metrics}
\resizebox{\textwidth}{!}{
\begin{tabular}{c|c|c|c|c|c|c|c|c|c}
\hline
\textbf{Perceptual Metric} & \textbf{Datasets} & \textbf{LDCT} & \textbf{BM3D} & \textbf{RED-CNN} & \textbf{CNCL} & \textbf{CTFormer} & \textbf{SwinIR} & \textbf{CoreDiff} & \textbf{PatchDenoiser} \\
\hline

\multirow{4}{*}{\textbf{SSIM}}
& \textbf{mayo2016ABD}  & 0.8521 & 0.8738 & 0.8967 & 0.8925 & 0.8853 & 0.9013 & 0.9033 & 0.8965 \\
& \textbf{mayo2020ABD}  & 0.8509 & 0.8799 & 0.9055 & 0.8968 & 0.8842 & 0.9047 & 0.9089 & 0.9035 \\
& \textbf{mayo2020CHEST} & 0.6757 & 0.6821 & 0.7264 & 0.7228 & 0.6891 & 0.7335 & 0.7369 & 0.7262 \\
& \textbf{piglet}       & 0.8581 & 0.8710 & 0.9300 & 0.9233 & 0.9022 & 0.9202 & NA & 0.9285 \\
\hline

\multirow{4}{*}{\textbf{FSIM}}
& \textbf{mayo2016ABD}  & 0.9436 & 0.9513 & 0.9582 & 0.9579 & 0.9587 & 0.9603 & 0.9634 & 0.9598 \\
& \textbf{mayo2020ABD}  & 0.9461 & 0.9557 & 0.9649 & 0.9619 & 0.9632 & 0.9640 & 0.9671 & 0.9657 \\
& \textbf{mayo2020CHEST} & 0.8442 & 0.8471 & 0.8743 & 0.8748 & 0.8748 & 0.8808 & 0.8911 & 0.8748 \\
& \textbf{piglet}       & 0.9440 & 0.9489 & 0.9690 & 0.9678 & 0.9586 & 0.9629 & NA & 0.9650 \\
\hline

\multirow{4}{*}{\textbf{NQM}}
& \textbf{mayo2016ABD}  & 28.3936 & 28.6598 & 29.5013 & 28.9024 & 28.8233 & 29.9116 & 29.9579 & 29.2352 \\
& \textbf{mayo2020ABD}  & 28.6407 & 28.9784 & 30.0049 & 29.0292 & 28.9935 & 29.7588 & 30.5737 & 29.4463 \\
& \textbf{mayo2020CHEST} & 12.2276 & 12.2170 & 18.3481 & 18.5280 & 14.7093 & 18.5272 & 19.6990 & 18.3582 \\
& \textbf{piglet}       & 23.6569 & 23.7755 & 27.6234 & 27.2361 & 25.5275 & 25.1146 & NA & 27.6408 \\
\hline

\end{tabular}}
\end{table*}

\begin{figure*}
\centering
\begin{subfigure}
    \centering
    \includegraphics[width=\textwidth]{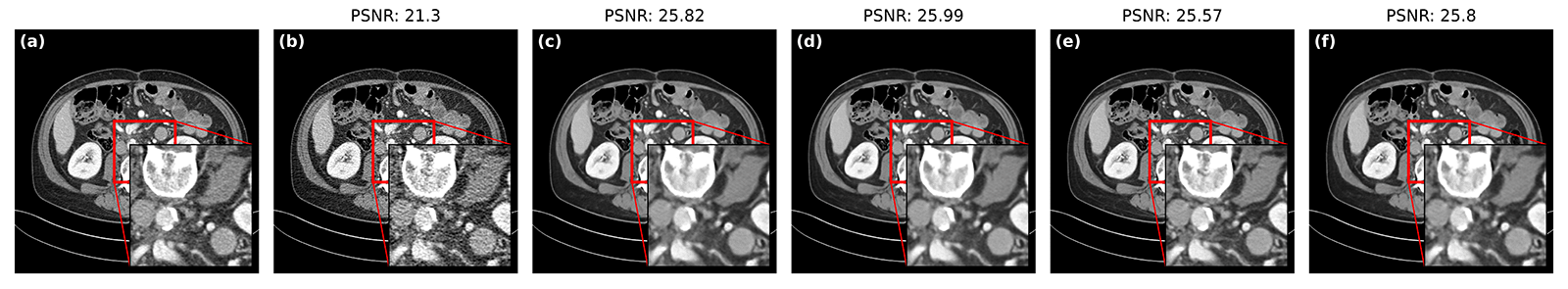}
\end{subfigure}
\begin{subfigure}
    \centering
    \includegraphics[width=\textwidth]{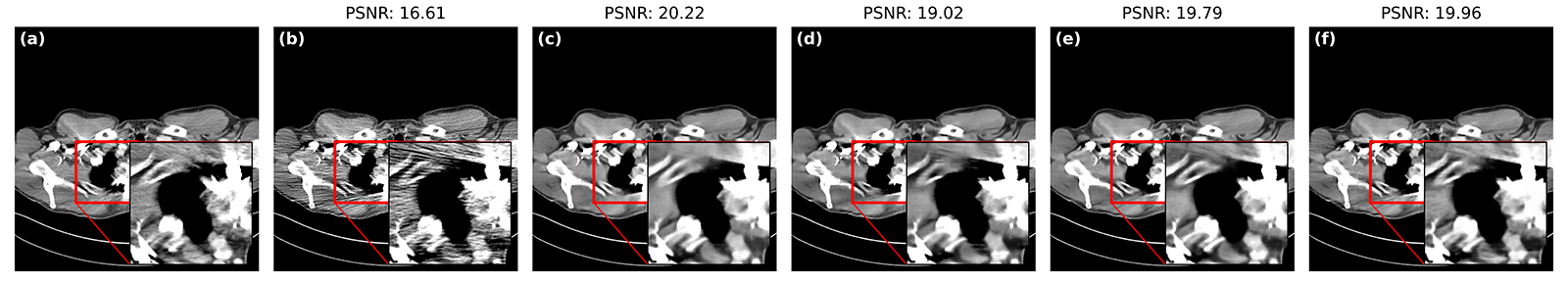}
\end{subfigure}
\begin{subfigure}
    \centering
    \includegraphics[width=\textwidth]{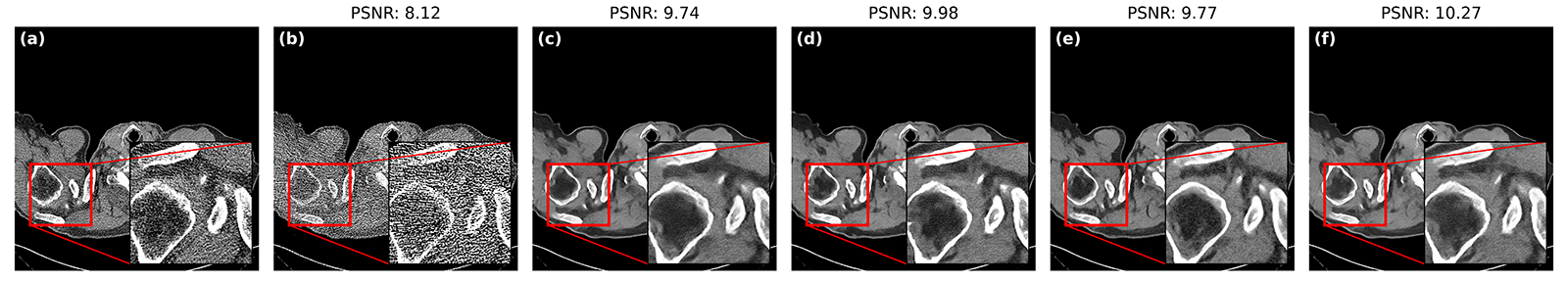}
\end{subfigure}
\begin{subfigure}
    \centering
    \includegraphics[width=\textwidth]{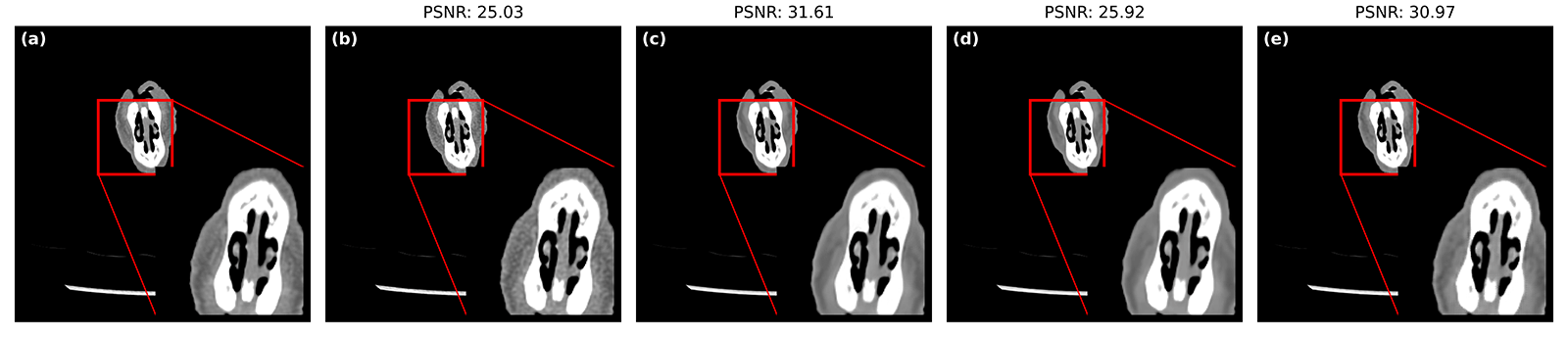}
\end{subfigure}
\caption{
Qualitative comparison of reconstructed CT images across four datasets: Mayo2016ABD, Mayo2020ABD, Mayo2020Chest, and Piglet. Each subfigure shows results for (a) NDCT, (b) LDCT, (c) RED-CNN, (d) SwinIR, (e) CoreDiff (except Piglet where CoreDiff is not available), and the proposed (f) PatchDenoiser. For each method, the corresponding PSNR values are overlaid on the images to provide quantitative reference alongside visual assessment.
}
\label{fig:qualitative_results}
\end{figure*}

Statistical significance was evaluated using the Wilcoxon signed-rank test with Holm correction on SSIM values for Mayo2016ABD. PatchDenoiser shows no statistically significant difference compared with RED-CNN after multiple-comparisons correction (adjusted $p=0.058$), indicating comparable structural preservation despite its lower computational complexity. PatchDenoiser significantly outperforms CNCL (adjusted $p<0.001$) with an average SSIM improvement of 0.004. In contrast, SwinIR and CoreDiff achieve statistically higher SSIM values than PatchDenoiser (adjusted $p<0.001$); however, these improvements are accompanied by increased model complexity. Overall, the results demonstrate that PatchDenoiser provides competitive reconstruction quality while offering a more efficient alternative to computationally intensive architectures.

\section{Cross Dataset Generalization}
To assess cross-dataset generalization, models trained on Mayo2020ABD were evaluated on unseen Mayo2016ABD and Mayo2020CHEST datasets without fine-tuning. PatchDenoiser achieves competitive performance under domain shift, obtaining 32.08 dB/0.8984 SSIM on Mayo2016ABD and 18.88 dB/0.7034 SSIM on Mayo2020CHEST. While advanced transformer-based methods achieve higher performance in some cases, PatchDenoiser consistently preserves structural information and outperforms several CNN-based approaches on the unseen chest dataset. These results demonstrate the robustness of PatchDenoiser across different anatomical regions and acquisition domains while maintaining its computational efficiency.

\begin{table}
\centering
\caption{Cross-dataset generalization performance. Models are trained on Mayo2020ABD and evaluated on unseen Mayo2016ABD and Mayo2020CHEST datasets without fine-tuning.}
\label{tab:cross_dataset}
\resizebox{\linewidth}{!}{
\begin{tabular}{lcccc}
\toprule
\multirow{2}{*}{Method} & \multicolumn{2}{c}{Mayo2016ABD} & \multicolumn{2}{c}{Mayo2020CHEST} \\
\cmidrule(lr){2-3} \cmidrule(lr){4-5}
& PSNR & SSIM & PSNR & SSIM \\
\midrule
RED-CNN       & 32.3469 & 0.8997 & 17.0057 & 0.6914 \\
CNCL          & 31.1310 & 0.8925 & 20.4806 & 0.7228 \\
CTFormer      & 30.8904 & 0.8853 & 19.5250 & 0.6891 \\
SwinIR        & 32.5143 & 0.9013 & 20.7321 & 0.7335 \\
CoreDiff      & 32.3816 & 0.9033 & 20.7722 & 0.7369 \\
PatchDenoiser & 32.0793 & 0.8984 & 18.8787 & 0.7034 \\
\bottomrule
\end{tabular}
}
\end{table}

\section{Ablation Studies}
We further conduct extensive ablation studies to analyze the contribution of each architectural component and to identify an optimal configuration for PatchDenoiser. Table \ref{tab:ablation} summarizes the results of these experiments. Specifically, we investigate the impact of patch-size selection, latent-feature dimensionality, loss-function design, fusion strategy (concatenation vs. gated fusion), learning-objective formulation (content vs. noise prediction), and the presence of the proposed Patch Consolidator module.

\begin{table}
\centering
\caption{Ablation study of PatchDenoiser. Baseline corresponds to the full proposed model. PSNR values are reported on Mayo2016ABD.}
\label{tab:ablation}
\begin{tabular}{l|l|c|c}
\hline
\textbf{Setting} & \textbf{Variant} & \textbf{PSNR} & \textbf{Remarks} \\
\hline

Baseline & -- & 32.0157 & Full proposed model \\
\hline

\multirow{2}{*}{A1} 
& A1.1 & 31.9214 & Patch size=(128,256,512) \\
& A1.2 & 31.9824 & Patch size=(32,64,512) \\
\hline

A2 & -- & 32.1836 & Latent dimension $\times$2 \\
\hline

A3 & -- & 31.8221 & Concat-based fusion \\
\hline

\multirow{2}{*}{A4}
& A4.1 & 32.1629 & L1 loss function \\
& A4.2 & 31.8077 & L2 + SSIM loss \\
\hline

A5 & -- & 25.0647 & Learning noise \\
A6 & -- & 31.8877 & without PCM \\
\hline

\end{tabular}
\end{table}

The results indicate that the choice of loss function has a measurable impact on reconstruction quality, with the L1 loss yielding a marginal improvement of approximately 0.15 dB in PSNR compared to alternative formulations; in contrast, directly learning the noise component rather than the reconstruction target results in substantial performance degradation, highlighting the importance of appropriate learning objectives for stable denoising.

We further observe that increasing the latent dimensionality yields a modest performance gain, consistent with expectations; however, the improvement remains limited, suggesting diminishing returns beyond a certain capacity. Additionally, variations in patch size negatively affect performance in both coarse (A1.1) and fine (A1.2) configurations, indicating that PatchDenoiser is sensitive to patch-scale selection and reinforcing the importance of balanced multi-scale design.

Overall, these ablation results validate the effectiveness of the proposed design choices and emphasize that PatchDenoiser achieves a favorable trade-off between model compactness and reconstruction quality. 

\section{Discussion}
The experimental results demonstrate that PatchDenoiser achieves decent denoising performance while maintaining high computational efficiency. This highlights an important direction in architectural design, as many recent medical image denoising models rely on large-scale deep networks with substantial computational and energy demands, which are increasingly less desirable from both economic and environmental perspectives. In contrast, PatchDenoiser provides a favorable trade-off by explicitly decomposing the denoising process into local texture modeling and global context aggregation, supported by an efficient multi-scale patching strategy and spatially aware feature fusion.

Unlike conventional encoder-based or patch-based architectures that progressively reduce spatial resolution and may discard fine structural details, PatchDenoiser preserves spatial dimensions throughout feature extraction. This design minimizes information loss during feature transformation and enables effective local-to-global interaction while maintaining spatial consistency in the reconstructed outputs.

\begin{figure}
    \centering
    \includegraphics[width=0.5\textwidth]{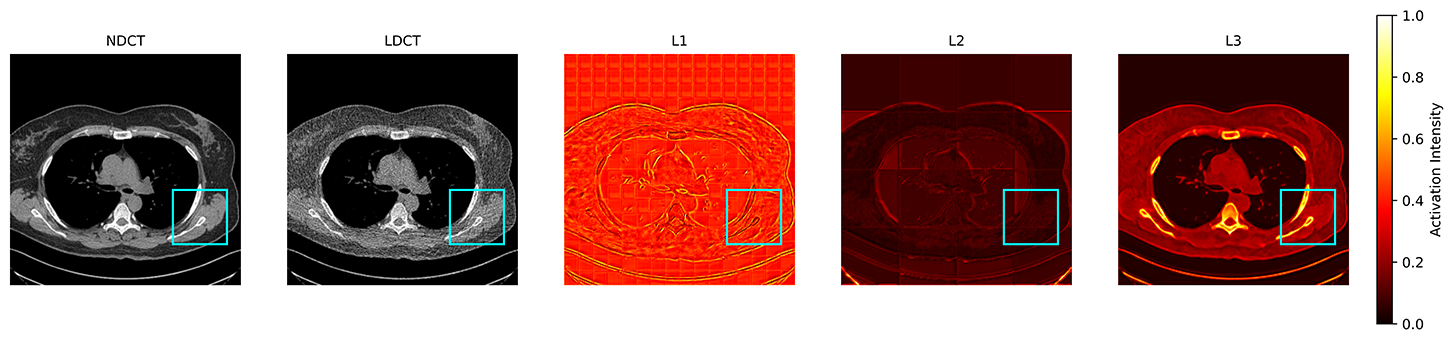}
    \caption{Visualization of intermediate feature representations extracted from different levels of the Patch Encoder.}
    \label{fig:levels-features}
\end{figure}

For interpretability, we analyze intermediate feature representations extracted from each Patch Encoder level, as illustrated in Figure \ref{fig:levels-features}. The visualization shows that higher-level features (L3) exhibit stronger responses to global anatomical structures, whereas lower-level features (L1) primarily focus on fine textures and local details. This decomposition indicates that each scale learns complementary representations of the underlying anatomy.

\begin{figure}
    \centering
    \includegraphics[width=0.5\textwidth]{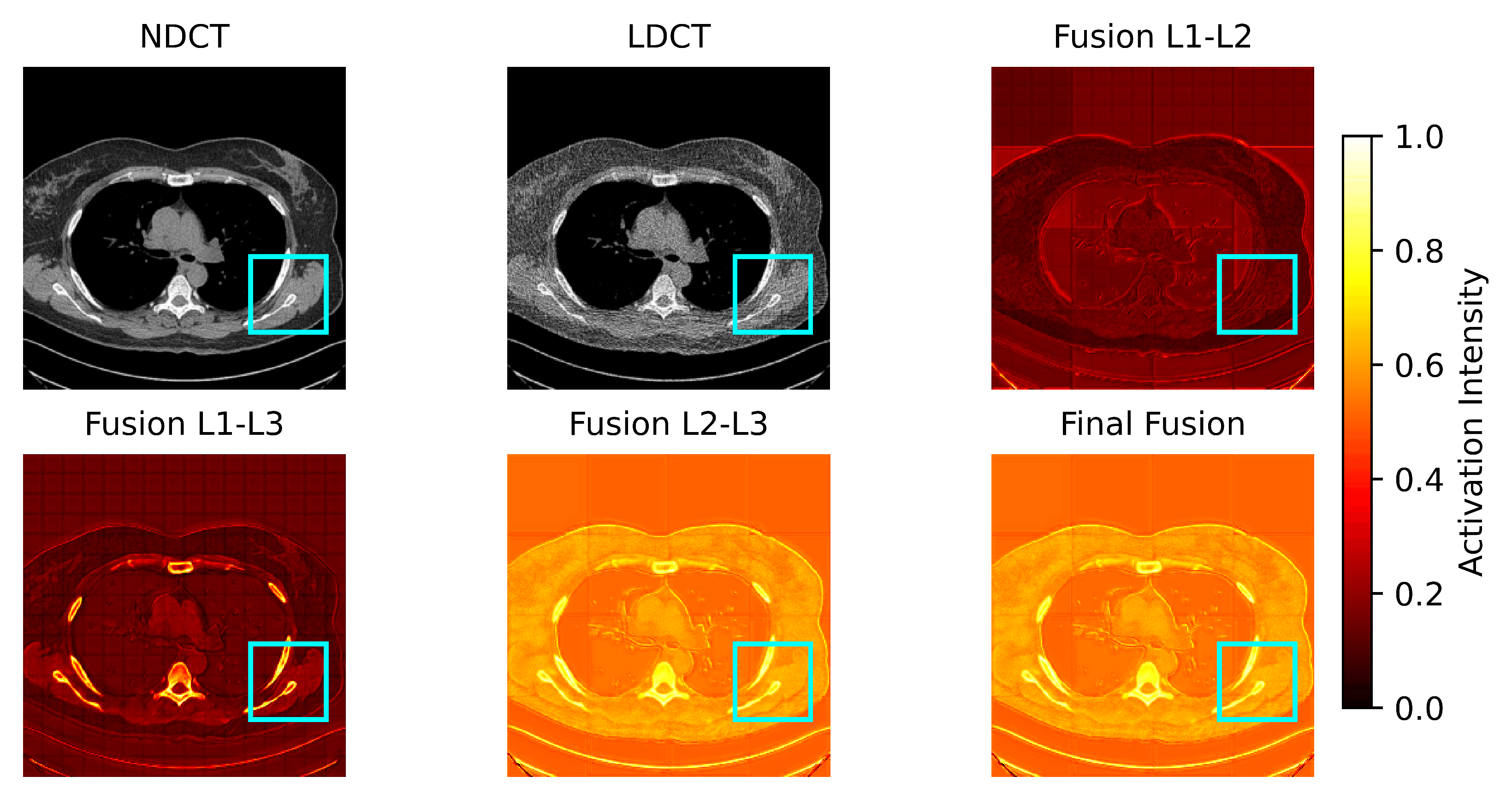}
    \caption{Feature maps after multi-scale fusion within the proposed PatchDenoiser. L1-L2 fusion integrates fine and mid-level information, while L1-L3 fusion emphasizes global contextual structures. L2-L3 fusion and the final fused representation show fusion of mid and high level features.}
    \label{fig:fused-features}
\end{figure}

Furthermore, Figure \ref{fig:fused-features} presents feature maps obtained after multi-scale fusion within the proposed architecture. The L1–L2 fusion captures a mixture of fine and mid-level structures, while L1–L3 fusion shows stronger responses to global contextual information, suggesting that the model adaptively emphasizes long-range structure during denoising. In contrast, L2–L3 fusion and the final fused representation exhibit more uniform activations, indicating effective integration of multi-scale features into a coherent reconstruction. Overall, these observations suggest that PatchDenoiser effectively combines complementary feature hierarchies, enabling robust reconstruction through spatially consistent fusion.

Despite these advantages, PatchDenoiser has certain limitations. The choice of patch size and the number of patch scales significantly influence performance. Although patch dimensions are adaptively derived from the input resolution, performance degrades when excessively small patches are used, likely due to insufficient contextual coverage. Similarly, the depth of the Patch Feature Extractor remains partially empirically determined, and a more systematic study is required to optimize this component fully.

In addition, patch boundary artifacts may arise during feature aggregation. In this work, the proposed Patch Consolidator Module effectively mitigates such effects; however, seamless patch integration remains a general challenge in patch-based architectures and warrants further investigation.

Moreover, PatchDenoiser is most effective in scenarios where local anatomical structures dominate image formation, while long-range global dependencies are less critical. Consequently, its applicability may be limited for tasks requiring extensive global reasoning. Nevertheless, in medical image denoising, where local texture preservation and noise suppression are primary objectives, this design choice is well aligned with the problem domain's intrinsic characteristics.

From a practical perspective, PatchDenoiser offers clear advantages in clinical deployment. While transformer- and GAN-based methods may occasionally achieve marginal improvements in PSNR or SSIM, they typically incur significantly higher computational cost and model complexity, often leading to diminishing returns in real-world settings. Furthermore, such large-scale models are often more sensitive to training data size and distribution shifts, which may limit their robustness in clinical environments.

Since denoising typically serves as a preprocessing step in broader medical imaging pipelines, lightweight and efficient architectures are particularly desirable. In this context, PatchDenoiser provides a balanced trade-off between reconstruction quality and computational efficiency, making it well suited for practical and scalable clinical applications.

\section{Conclusion}
In this work, we presented PatchDenoiser, a parameter- and energy-efficient multi-scale patch-based framework for medical image denoising. By explicitly separating local texture extraction from global context aggregation and employing spatially aware feature fusion, the proposed model achieves comparable denoising performance w.r.t to SOTA CNN-based methods while maintaining an ultra-lightweight architecture. 

To the best of our knowledge, PatchDenoiser is among the most computationally efficient LDCT denoising frameworks to date, offering a favorable balance between denoising performance and computational cost. These characteristics make it particularly well suited for integration into practical clinical AI pipelines.
Future work will focus on further enhancing the adaptability and generalization of the proposed framework, including the development of dynamic patch selection strategies and the incorporation of mechanisms to capture long-range global dependencies better. Extending PatchDenoiser to additional imaging modalities and integrating it into end-to-end clinical AI pipelines also represent promising directions. Furthermore, exploring model compression and quantization techniques may further improve energy efficiency and facilitate deployment in real-world clinical environments. 

\printbibliography

\begin{table*}
\centering
\caption{\textbf{Training configuration and convergence details across datasets and methods.} Val freq denotes validation frequency, and early stopping iteration corresponds to the iteration where training was terminated based on validation PSNR convergence criteria.}
\small
\setlength{\tabcolsep}{5pt}
\renewcommand{\arraystretch}{1.2}
\begin{tabular}{c|c|c|c|c}
\hline
Dataset & Model & Train iters & Val freq & Early stopping iter \\
\hline

\multirow{6}{*}{mayo2016ABD} 
& REDCNN        & 10000  & 1000 & 4846 \\
& CNCL          & 20000  & 1000 & 7720 \\
& CTFormer      & 30000  & 2000 & 27754 \\
& SwinIR        & 50000  & 5000 & 24562 \\
& CoreDiff      & 80000  & 5000 & 40000 \\
& PatchDenoiser & 10000  & 1000 & 8722 \\
\hline

\multirow{6}{*}{mayo2020ABD} 
& REDCNN        & 12000  & 1000 & 6000 \\
& CNCL          & 30000  & 1000 & 12000 \\
& CTFormer      & 40000  & 2000 & 18000 \\
& SwinIR        & 60000  & 5000 & 23438 \\
& CoreDiff      & 140000 & 5000 & 65000 \\
& PatchDenoiser & 12000  & 1000 & 10000 \\
\hline

\multirow{6}{*}{mayo2020CHEST} 
& REDCNN        & 12000  & 1000 & 6000 \\
& CNCL          & 30000  & 1000 & 12000 \\
& CTFormer      & 40000  & 2000 & 14000 \\
& SwinIR        & 60000  & 5000 & 15000 \\
& CoreDiff      & 140000 & 5000 & 40000 \\
& PatchDenoiser & 12000  & 1000 & 6000 \\
\hline

\multirow{5}{*}{piglet} 
& REDCNN        & 10000  & 1000 & 8922 \\
& CNCL          & 20000  & 1000 & 7586 \\
& CTFormer      & 30000  & 2000 & 30000 \\
& SwinIR        & 50000  & 5000 & 30398 \\
& PatchDenoiser & 10000  & 1000 & 9914 \\
\hline

\end{tabular}
\label{tab:training_details}
\end{table*}

\end{document}